\documentclass[letterpaper, 10 pt, conference]{ieeeconf}  

\IEEEoverridecommandlockouts                              

\usepackage{graphicx} 
\usepackage{caption}
\usepackage{subcaption}
\usepackage{amsmath}
\usepackage{url} 
\usepackage{comment}
\usepackage{booktabs}
\usepackage{cite}

\title{\LARGE \bf
TactEx: An Explainable Multimodal Robotic Interaction Framework for Human-Like Touch and Hardness Estimation
}

\author{Felix Verstraete*, Lan Wei*, Wen Fan*, and Dandan Zhang
\thanks{*Equal Contribution.}
\thanks{Felix Verstraete, Lan Wei, Wen Fan and Dandan Zhang are with the Department of Bioengineering, Imperial-X Initiative, Imperial College London, London, United Kingdom.  Corresponding: d.zhang17@imperial.ac.uk.}
}

\begin{document}
\thispagestyle{plain}

\maketitle

\begin{abstract}

Accurate perception of object hardness is essential for safe and dexterous contact-rich robotic manipulation. Here, we present TactEx, an explainable multimodal robotic interaction framework that unifies vision, touch, and language for human-like hardness estimation and interactive guidance. 
We evaluate TactEx on fruit-ripeness assessment, a representative task that requires both tactile sensing and contextual understanding. The system fuses GelSight-Mini tactile streams with RGB observations and language prompts. A ResNet50+LSTM model estimates hardness from sequential tactile data, while a cross-modal alignment module combines visual cues with guidance from a large language model (LLM). This explainable multimodal interface allows users to distinguish ripeness levels with statistically significant class separation (p $<$ 0.01 for all fruit pairs).
For touch placement, we compare YOLO with Grounded-SAM (GSAM) and find GSAM to be more robust for fine-grained segmentation and contact-site selection. A lightweight LLM parses user instructions and produces grounded natural-language explanations linked to the tactile outputs. In end-to-end evaluations, TactEx attains 90\% task success on simple user queries and generalises to novel tasks without large-scale tuning. These results highlight the promise of combining pretrained visual and tactile models with language grounding to advance explainable, human-like touch perception and decision-making in robotics.

\end{abstract}

\section{INTRODUCTION}

Humans rely on touch to infer intrinsic physical properties (e.g., hardness, friction, compliance) and to regulate contact, especially when vision is unreliable due to occlusion, poor lighting, or visually ambiguous materials \cite{su2012use}. In contact-rich interaction, mechanical properties govern how objects deform under applied forces and whether contact remains stable. Touch therefore enables not only stable manipulation but also interpretable comparative judgments (e.g., “this fruit is softer than that one”), which are often ambiguous from visual cues alone.

In robotics, tactile sensing provides complementary information that cannot be directly recovered from vision. High-resolution tactile imaging supports texture recognition \cite{zhang2025design} and hardness estimation \cite{Yuan2017}, while high-bandwidth tactile cues enable early slip detection and precise force modulation \cite{li2018slip}. These capabilities make tactile sensing particularly important for intrinsic property inference during physical interaction.
Among the intrinsic properties accessible through touch, hardness is especially critical for manipulation. Accurate hardness estimation supports safe force regulation, adaptive grasping, and delicate object handling. However, fine-grained hardness discrimination requires controlled contact, high-resolution tactile sensing, and statistically reliable comparison, particularly when visually similar objects differ mechanically. Fruit ripeness assessment exemplifies this challenge: subtle compliance variations must be distinguished under visually ambiguous conditions \cite{Liao2025}. Beyond its practical relevance in agriculture and domestic robotics, ripeness assessment provides a principled benchmark for multimodal hardness estimation.

Despite recent advances in tactile hardware \cite{fan2025crystaltac} and tactile-driven manipulation, most approaches prioritize task success \cite{10160288}, such as pushing, grasping, or tactile servoing ,rather than explicit intrinsic property inference. Hardness estimation models remain limited in practice, particularly for fine-grained ripeness discrimination. Existing methods often require large datasets or extensive fine-tuning \cite{Fu2024, Yu2024, Nam2024}, and many lack statistical validation for subtle compliance differences \cite{Chen2025}. As a result, calibrated and reliable hardness estimation remains an open challenge.

Beyond accuracy, explainability is critical for deploying tactile perception in human-facing applications. When robots estimate intrinsic properties, they should be able to justify how tactile evidence supports the inference and how that inference informs subsequent control decisions. While structured tactile representations \cite{fan2022graph} and interpretable policy learning \cite{zhang2021explainable} improve transparency at the feature or policy level, they rarely connect calibrated property estimation to operator-facing explanations grounded in tactile measurements. In line with Adebayo et al.’s xAI framework for robotics \cite{adebayo} and prior work emphasizing user-oriented explanations \cite{SETCHI20203057}, practical systems should provide transparent reasoning, uncertainty reporting, and sensor-grounded rationales. Without these capabilities, trust and adoption in human-facing settings remain limited.
These gaps motivate a framework that treats hardness estimation as a first-class perceptual objective within a structured, multimodal, and explainable interaction loop.

To address this need, we propose \textbf{TactEx}, an explainable multimodal robotic interaction framework for touch-based hardness estimation (Fig.~\ref{fig:overview}). TactEx integrates language grounding, visual perception, controlled tactile exploration, and statistically grounded hardness inference in a modular pipeline. Given a natural-language query, the robot localizes the referenced objects, performs top-down probing with a high-resolution tactile sensor, and estimates hardness from the resulting contact sequence. A large language model (LLM) then produces concise, sensor-grounded explanations and comparisons, such as ripeness or hardness rankings across the queried fruits.

TactEx is designed around three principles: (i) multimodal interaction, where touch is treated as a primary perceptual signal coupled with exploratory contact; (ii) human-like hardness reasoning, enabling fine-grained comparative estimation with confidence reporting; and (iii) explainability, providing intermediate outputs and uncertainty-aware rationales grounded in tactile evidence. Together, these components support accurate and interpretable hardness estimation in contact-rich, human-facing settings.

This paper makes the following contributions:
\begin{enumerate}
    \item \textbf{Data‑efficient visuo‑tactile hardness regression.} We introduce a pretrained ResNet50 + LSTM hardness estimation pipeline that achieves RMSE 4.3 and $\rho=0.88$ using only N=280 fine-tuning samples, with statistically significant ripeness ranking across five fruit types.

    \item \textbf{Language-conditioned servoing for touch perception.} We demonstrate that text-prompted Grounded-SAM outperforms YOLO in touch placement accuracy, leading to improved downstream hardness estimation reliability.

    \item \textbf{End-to-end explainable multimodal interaction.}  We present a language-grounded framework that converts user queries into visuo-tactile actions and sensor-grounded explanations, achieving high object- and scenario-level success across four interaction complexity tiers.
\end{enumerate}

\section{Related Work}
\subsection{Hardness and Ripeness Estimation}
While several works treat hardness as a classification problem using GelSight or traditional tactile sensors \cite{Guo2025, Chen2025}, discrete labels lack the resolution required for objects that cluster tightly at the high end of the Shore 00 scale. Because many fruits and vegetables fall within this narrow band, classification systems often fail to distinguish subtle mechanical variations, such as the difference between an unripe and a ripe banana. This limitation necessitates a regression formulation capable of resolving these fine-grained compliance differences.

Physics and feature-engineered methods aim to estimate hardness by tracking physical changes in tactile images or force signals. Yuan et al. proposed a numerical model based on the changing brightness in the tactile image and the  force variations \cite{Yuan2024}, and Liao et al. leveraged force dynamics for ripeness tracking \cite{Liao2025}. Although effective in controlled settings, these approaches rely on hand-crafted features, impose shape or pose constraints, show limited generalizability across objects and contact regimes, and rarely integrate other modalities.

Deep learning approaches, such as the widely adopted VGG16-LSTM GelSight baseline by Yuan et al. \cite{Yuan2017}, have demonstrated strong performance in foundational robotic applications. However, these unimodal systems do not report statistically validated rankings for fine-grained properties like ripeness, restricting their real-world applicability. Similarly, recent work by Nam et al. \cite{Nam2024} explored continuous tactile regression but encountered performance degradation when assessing harder objects (upper Shore 00 range) and did not incorporate an explainable, language-conditioned interface.

\subsection{Language Grounded Multimodal Models}
Recent work increasingly integrates language with vision and tactile sensing. In these systems, language is used to interpret user requests and to generate responses \cite{Guo2025, Ueda2024, Zhao2023}. For example, Ueda et al. demonstrate that a robot equipped with tactile sensors can leverage the zero-shot capabilities of vision–language models to recognize objects. 
Tactile-VLA shows that augmenting vision–language models with touch improves the translation of user intent into precise physical actions \cite{huang2025tactilevlaunlockingvisionlanguageactionmodels}. Force-VLA reports performance gains in contact-rich manipulation \cite{yu2025forcevlaenhancingvlamodels}, and VTLA further improves challenging insertion tasks \cite{zhang2025vtlavisiontactilelanguageactionmodelpreference}. Despite these advances, the focus remains primarily on action execution, with less attention to improving perception and language-conditioned reasoning about object properties.
In this study, we plan to integrate language and vision to support perception by helping to locate and filter objects in the scene.

\begin{figure*}[!h]
    \centering
    \includegraphics[width=1\linewidth]{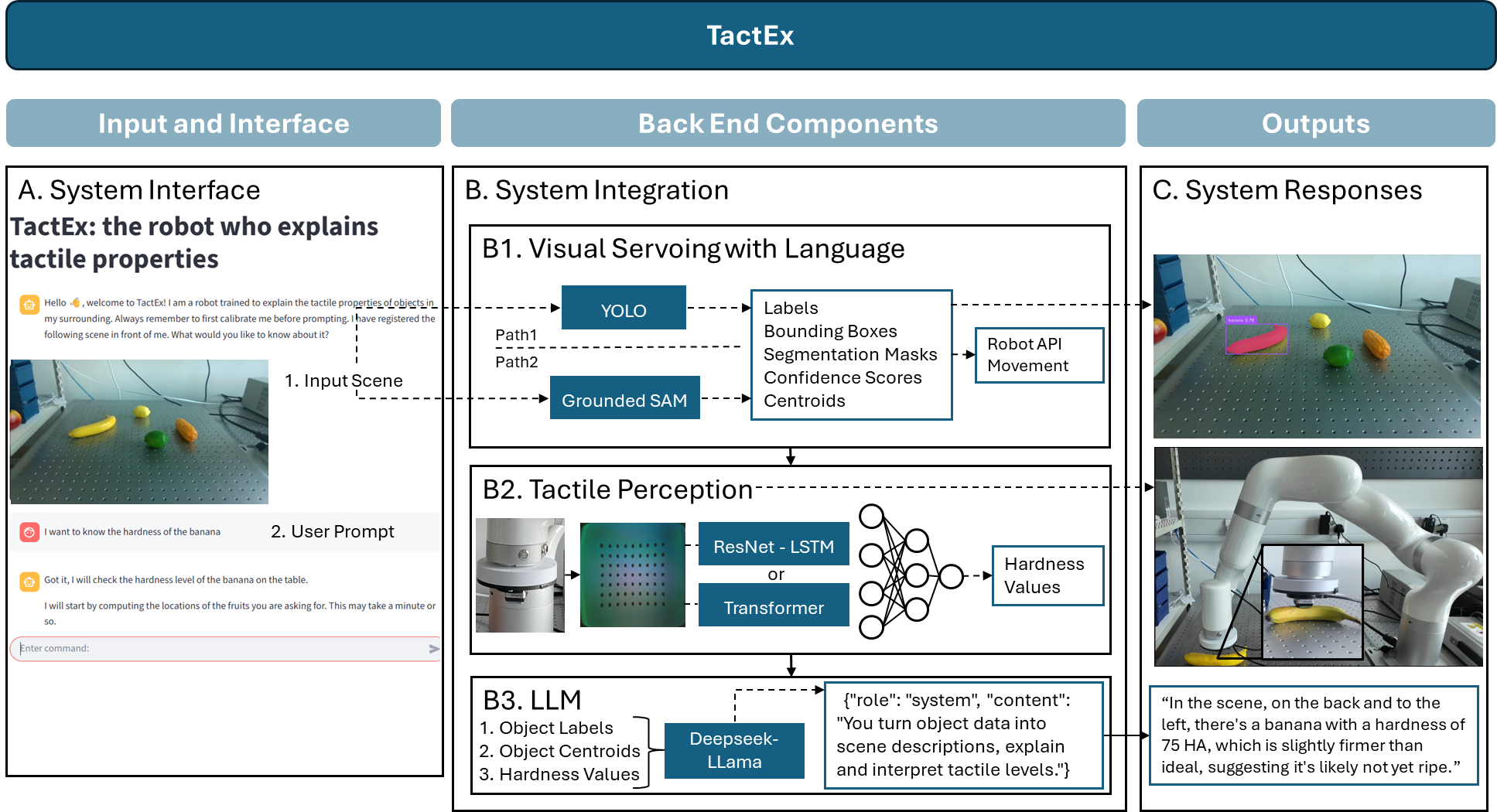}
    \captionsetup{font=footnotesize,labelsep=period}
    \caption{Overview of TactEx (“The Tactile Explainer”), a multimodal framework for fruit ripeness explanation. Users interact via a chat interface (A), objects are localized with YOLO or GSAM (B1), hardness is estimated with a GelSight sensor (B2), and an LLM composes the final response from the fruit names, locations and hardness values (B3–C). The three components are detailed in \ref{sec:serv}, section \ref{sec:tac} and section \ref{sec:llm}, respectively.}
    \label{fig:overview}
    \vspace{-0.4cm}
\end{figure*}

\section{Methods}

We present \textbf{TactEx} (“The Tactile Explainer”), a modular and explainable multimodal robotic framework that \emph{perceives} and \emph{verbally explains} object hardness and fruit ripeness (Fig.~\ref{fig:overview}). 
The system comprises three stages: (A) \textit{Input and Interface}, (B) \textit{Back-end Processing}, and (C) \textit{Outputs}.

\subsection{System Interface}

The user interface (Fig.~\ref{fig:overview}, A) is implemented in Streamlit, providing a lightweight and interactive front end for human–robot interaction. 
Upon launch, the interface displays a live RGB stream from the robot-mounted camera, showing fruits and vegetables arranged on a workspace. 
A chat window below the video feed accepts free-form natural-language prompts (e.g., “Which fruit is the softest?” or “How ripe are the banana and the lemon?”).

Submitting a query initiates the back-end processing pipeline (Fig.~\ref{fig:overview}, B). 
The system first interprets the user’s language instruction to determine the relevant objects. 
It then executes perception, manipulation, tactile inference, and language generation in sequence, as described below.

\paragraph{Vision and Object Grounding}
We employ the Grounded Segment-Anything Model (GSAM) for zero-shot, text-promptable object identification in unstructured household environments \cite{TianheRen2024}. 
Conditioned on the user’s language query, GSAM returns instance segmentation masks and corresponding object locations (Fig.~\ref{fig:overview}, B1). 
We use these outputs to initialise and guide visual servoing, allowing accurate end-effector alignment and positioning for subsequent tactile probing.
We compare GSAM against a conventional baseline that combines YOLO-based object detection with language-conditioned selection \cite{Bai2020}. 
In this pipeline, YOLO provides candidate bounding-box detections, while the language instruction is used to select the subset of detections matching the query.


\paragraph{Tactile Acquisition and Hardness Estimation}
Once the target objects are localised, the robot performs a controlled top-down pressing motion using a GelSight-Mini tactile sensor to capture a sequence of contact images. 
The resulting spatiotemporal tactile data are processed by either a ResNet–LSTM architecture or a Transformer-based regressor to estimate object hardness (Fig.~\ref{fig:overview}, B2).

\paragraph{Language Generation and Output}
Finally, the estimated hardness values are combined with object labels and spatial context to produce a concise natural-language explanation via a large language model (Fig.~\ref{fig:overview}, B3). 
The interface then presents the results to the user (Fig.~\ref{fig:overview}, C), including object positions, predicted hardness values, and inferred ripeness expressed in user-friendly terms.

\subsection{System Components}
\subsubsection{Visual Servoing Methods}
\label{sec:serv}

We employ an eye-to-hand configuration for visual servoing. An Intel RealSense depth camera is mounted above the workspace, with intrinsic parameters $f_x=608.5$, $f_y=606.9$, $c_x=309.4$, and $c_y=213.83$. 
Workspace coordinates are expressed in metric units (mm) by projecting pixel detections into 3D space using camera intrinsics and extrinsics obtained via calibration with a 7×5 ChArUco checkerboard (square size: 2.5 cm; marker size: 1.8 cm).

The user’s request is first parsed using a lightweight NLP module to extract the target fruit classes and intended comparison (e.g., hardness ranking). These extracted object labels condition the subsequent perception stage.
We compare two segmentation approaches: (i) a YOLOv8-based instance segmentation model and (ii) the Grounded Segment-Anything Model (GSAM) \cite{TianheRen2024}.

The YOLOv8 model was trained on a custom dataset of 60 images containing up to 12 fruits and vegetables per scene for 100 epochs. Ground-truth annotations were manually generated using Roboflow. Training was performed with default hyperparameters (pretrained=True, learning rate=0.01, optimizer=SGD). At inference time, a confidence threshold of 0.40 was selected to balance false positives and missed detections. Since YOLO predicts all detectable object classes, NLP-based filtering is applied to retain only those objects referenced in the user’s query.
In contrast, GSAM requires no task-specific training. A higher confidence threshold (0.60) is applied during inference to reduce spurious segmentations. NLP filtering is likewise used to remove objects not mentioned in the request.

\begin{figure}[!t]
    \begin{subfigure}[b]{0.156\textwidth}
        \centering        
        \includegraphics[width=\textwidth]{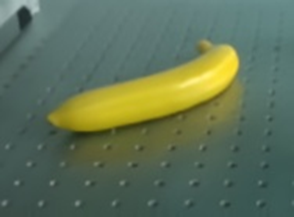}
        \caption{}
        \label{fig:sub1}
    \end{subfigure}
    \begin{subfigure}[b]{0.156\textwidth}
        \centering
        \includegraphics[width=\textwidth]{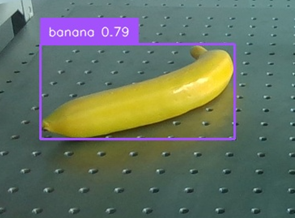}
        \caption{}
        \label{fig:sub2}
    \end{subfigure}
    \begin{subfigure}[b]{0.156\textwidth}
        \centering
        \includegraphics[width=\textwidth]{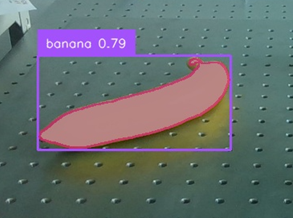}
        \caption{}
        \label{fig:sub4}
    \end{subfigure}
    \captionsetup{font=footnotesize,labelsep=period}
    \caption{Example of the Grounded SAM procedure: (a) original scene, (b) object detection with bounding box, (c) results SAM with inner mask for computing the centroid.}
    \label{fig:all_subfigures}
        \vspace{-0.4cm}
\end{figure}

To mitigate the effect of noisy depth measurements on 3D localisation, both pipelines employ a multi-stage robustness procedure:
(i) For GSAM, Grounding DINO first predicts a bounding box from the language prompt (Fig.~\ref{fig:sub2}).  
(ii) Instance masks are then obtained from YOLO or GSAM.  
(iii) To suppress boundary artefacts, mask edges are detected using a Canny operator and dilated with a 3×3 kernel (two iterations), producing an inner mask that excludes noisy contours (Fig.~\ref{fig:sub4}).  
(iv) Depth at each pixel within the refined mask is stabilised by computing the median across 10 consecutive frames.  
(v) The 3D centroid is finally computed as the median of the filtered mask coordinates projected into metric space.
This procedure improves localisation stability and reduces the influence of depth outliers on visual servoing performance.

\subsubsection{Tactile Perception Methods}
\label{sec:tac}

We compare three tactile perception models for hardness estimation. Each model combines (i) a CNN backbone for per-frame feature extraction from GelSight RGB images and (ii) a temporal encoder to aggregate information across the contact sequence.

We adopt a VGG16--LSTM architecture as a baseline, following early work on vision-based tactile perception \cite{Yuan2017}. The CNN encodes each tactile frame into a feature vector, which is then integrated over time by an LSTM to predict hardness.
Motivated by evidence that residual networks improve representation quality in tactile tasks (e.g., slip detection and grasp outcome prediction), we replace VGG16 with a ResNet50 backbone while keeping the LSTM temporal module fixed \cite{Calandra2017, Gao2023}. This isolates the effect of the CNN backbone on hardness estimation performance.
Finally, we evaluate a non-recurrent temporal encoder by pairing a ConvNeXt backbone with a Transformer. In contrast to LSTMs, Transformers model long-range dependencies via self-attention over the entire sequence, potentially improving robustness to subtle temporal cues during pressing.

Each model takes as input a sequence of $T\in\{2,4\}$ RGB tactile images resized to $224\times224$. Table~\ref{tab:hyper} summarises the principal hyperparameters. For LSTM-based models, we use three stacked LSTM layers.

\begin{table}[!t]
\centering
\setlength{\tabcolsep}{4pt} 
\captionsetup{font=footnotesize,labelsep=period}
\caption{Model hyperparameters for base models. CNN: CNN output; LSTM/TF: hidden dim of LSTM or Transformer; nH,FF: attention heads / feedforward dimension; FC/Out: fully connected layers / output.}
\label{tab:hyper}
\begin{tabular}{lcccccc}
\toprule
Model & CNN & LSTM/TF & nH & FF & FC/Out \\
\midrule
VGG16-LSTM & 4096 & 512 & - & - & 256→128→32→1 \\
ResNet50-LSTM & 2048 & 512 & - & - & 256→128→32→1 \\
ConvNext-TF & 256 & 256 & 4 & 512 & 128→1 \\
\bottomrule
\end{tabular}
    \vspace{-0.6cm}
\end{table}

To avoid model collapsing, which means that all predictions are mapped to the mean of the training set, three extra measures in model training were taken. 
First, dropout layers (factor 0.1 to 0.2) were added to the LSTM and fully connected layers. Additionally, the AdamW optimiser used a higher learning rate in later layers (1e-3) in comparison with early layers (5e-5). Finally, a custom loss function (equation \ref{eq:loss}) was constructed based on the mean squared error (MSE) between the predictions ($p$) and the labels ($l$) and a penalty for low variability (Var) within the predictions:

\begin{equation}
\label{eq:loss}
\mathcal{L} = \text{MSE}(p, l)+4\min \Bigg( \frac{1}{\text{Var}({p})+10^{-6}}, 1000 \Bigg)
\end{equation}

\subsubsection{LLM}
\label{sec:llm} 
For generating responses to user requests, we employed a DeepSeek-R1-Distill-Llama-70B model, accessed via Groq. The model was configured with a temperature of 0.1 to reduce hallucination while keeping natural flow. The system was assigned the following role: \textit{``You turn object data into scene descriptions, explain and interpret tactile levels.''} The prompt included 10 rules, covering different aspects for a more precise response. These included, but were not limited to: (i) object location descriptions translated from workspace coordinates via a rule-based mapping (e.g., left/center/right, front/center/back); (ii) ideal ripeness thresholds for bananas, limes, and lemons (defined empirically by comparing ripe and unripe fruits to reference hardness objects); and (iii) writing style guidance to ensure concise, fluent, and operator-friendly language.

\section{Experiments}

\subsection{Dataset Description}
\subsubsection{\textbf{Visual Servoing Dataset}} 
We evaluate in tabletop scenes containing 1 to 5 fruits. The dataset includes 40 annotated instances (4 per each of 10 fruit types) with bounding boxes, segmentation masks, and centroids, enabling detection, segmentation, and centroid accuracy assessment.

\subsubsection{\textbf{Tactile Perception Dataset}} 
\label{sec:datacoll}
We use three splits: (i) a pretraining set drawn from an online GelSight dataset of approximately 5{,}000 objects spanning the Shore 00 scale, where contact frames are detected at $\mathrm{SSIM} < 0.90$ relative to a no-contact reference and 8-frame clips are extracted~\cite{Yuan2017,Zhang2024}; (ii) a fine-tuning set collected with a marker-based GelSight-Mini mounted on a uFactory 850 robotic arm; and (iii) a validation set collected with the same setup. All collected sequences contain 8 images with 0.25\,mm inter-frame steps. For our collected data, contact is defined by $\mathrm{SSIM} \le 0.96$ and mean marker displacement $> 2$ pixels~\cite{Nam2024, s20133796}. The difference in contact criteria across datasets reflects minor sensor and illumination differences.

The fine-tuning objects are five rubber cubes (66 to 80 HA), an elastic band (88 HA), and a glasses pouch (62 HA). This selection covers most of the expected fruit hardness range (60 to 90 HA). For each object, we record 40 poses by varying $x$–$y$ position by $\pm 5$\,mm and yaw by 0 to $45^\circ$, yielding 280 samples in total.  
The validation set includes three fruit pairs (mango, lime, tomato) and two fruit trios (banana, avocado) at distinct ripeness stages. For each individual fruit, we collect 20 samples.

\begin{figure}[!t]
    \centering
    \captionsetup{font=footnotesize,labelsep=period}
    \includegraphics[width=1\linewidth]{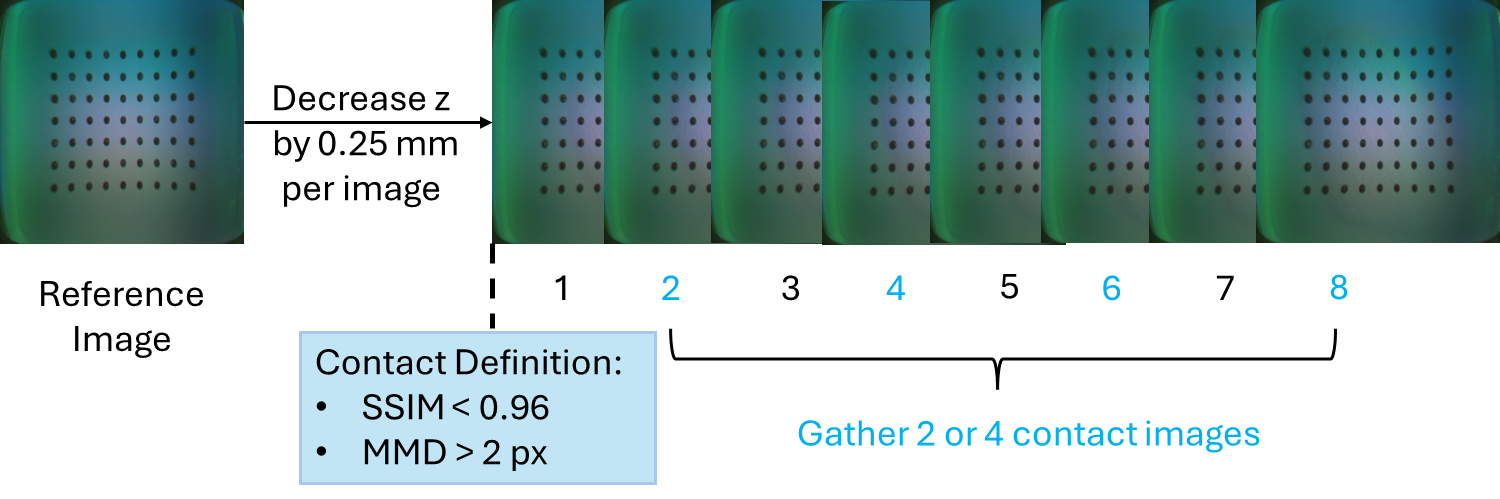}
    \caption{Data collection: images were compared to a reference image. If the contact criteria were met, 8 images were captured and transformed into a 2 or 4 image sequence.}
    \vspace{-0.3cm}
    \label{fig:datacollection}
\end{figure}

We convert each 8-frame contact sequence into shorter image sequences. For 2-frame sequences we use the 2nd and 8th frames; for 4-frame sequences we additionally include the 4th and 6th frames (Fig.~\ref{fig:datacollection}). These selections capture early and peak deformation while maintaining coverage of the contact trajectory. Finally, the difference between the selected images and the first contact image was used as input.

\subsubsection{\textbf{LLM Dataset}} 
We construct an evaluation set of 100 prompts based on randomly generated scenes featuring 1 to 6 fruits. Each prompt references objects present in the validation tactile set so that language outputs can be grounded in measured hardness. Prompts span common intents, including locate-and-measure requests, pairwise or list-wise comparisons, ranking by hardness or ripeness, and filtering by attributes such as color or position. For each prompt, the expected output includes referenced object labels, per-object hardness estimates or intervals and an ordered list when applicable.

\subsection{Experimental Setup}
\subsubsection{\textbf{Visual Servoing Setup}} 
Each scene is queried through the interface with the prompt
\emph{“I want to know the hardness of the [fruit]”}, which triggers detection and centroid computation as described in Sec.~\ref{sec:serv}.

\subsubsection{\textbf{Tactile Perception Setup}} 
\label{sec:training strategy}
Backbones are initialised from ImageNet, then pretrained (80 epochs) on the online data and fine-tuned (15 epochs) on collected data. 

All images were augmented by horizontal flips and colour jitter to increase model robustness to orientation variations and lighting conditions. The latter employed a brightness, contrast and saturation variation of 10\% and hue by 1\%. Furthermore, a cosine LR scheduler (patience 2, factor 0.2), weight decay of 1e-4 and batch size of 8 were used. 
 
\subsubsection{\textbf{LLM Setup}} 
An LLM-as-a-judge is used to evaluate the prompts. Within this setup, a dedicated LLM model (Llama-4-Maverick-17b-128e-Instruct) is instructed to score the prompt from 1 to 5 based on three metrics \cite{Gu2025}.

\subsubsection{\textbf{Complete Framework Setup}} 
\label{sec:completeframe}

The complete framework is evaluated based on four scenarios varying in complexity \cite{Verbaan2024, Calandra2017}. 
The complexity level is determined by three factors: 1) number of objects prompted , 2) number of distinct objects requested and 3) language reasoning. The latter one is determined by whether the fruits are explicitly mentioned in the request. Overall complexity is then categorised as low, medium and high, as shown in Table \ref{tab:complexity}. 

\begin{table}[!t]
    \centering
    \captionsetup{font=footnotesize,labelsep=period}
    \caption{Scenario complexity breakdown. [object] = chosen items, [property] = hardness, ripeness or softness. Complexity is determined by the number of objects, their distinctness, and whether fruits are explicit.}
    \renewcommand{\arraystretch}{1.2}
    \setlength{\tabcolsep}{2pt} 
    \resizebox{\columnwidth}{!}{%
    \begin{tabular}{|c|p{4.5cm}|c|c|c|c|}
        \hline
        \textbf{ID} & \textbf{Prompt} & \textbf{\# Obj.} & \textbf{\# Distinct} & \textbf{Explicit?} & \textbf{Complexity} \\
        \hline
        1 & Identify the [property] of [object]. & 1 & 1  & Yes & Low \\
        \hline
        2 & Identify the most [property] [object] in the scene. & 2 & 1 & Yes & Medium \\
        \hline
        3 & Summarize the [property] of the [object], [object] and [object]. & 3 & 3 & Yes & Med-High \\
        \hline
        4 & Summarize the [property] of all fruits in the scene. & 5 & 3–5 & No & High \\
        \hline
    \end{tabular}}
    \label{tab:complexity}
    \vspace{-0.6cm}
\end{table}

\subsection{Evaluation Metrics}
\subsubsection{\textbf{Visual Servoing Metrics}} 

Both models are evaluated using three metrics: confidence score, segmentation score (intersection over union or IoU) and the distance between the ideal midline of the fruit and the computed centroid. The ground-truth masks were manually annotated using Roboflow. These three metrics were compared using an independent t-test with unequal variances ($\alpha=0.01$), as Levene's test revealed p-values below 0.01. The null hypothesis of these t-tests is that there is no difference in the means between YOLO and GSAM. Although normality (Shapiro-Wilk) was not met for all groups, we assume the central limit theorem holds true given the large sample size (40).

Additionally, the errors were compared using a one-sample t-test ($\alpha=0.01$) to a threshold of 5mm to test if tactile perception would be valid. This threshold was indeed the deviation from the centre we allowed during data collection of the tactile models (see section \ref{sec:datacoll}). Finally, also the success rate (SR) was noted. In this case, success is defined as the model hoovering toward the correct fruit.

\subsubsection{\textbf{Tactile Perception Metrics}}

 The test set for pretraining comprised 20\% of the online data (N=962). For fine-tuning, the model run 7 times with a leave-one-out procedure across the object. In order to select the most optimal model after pretraining and fine-tuning, the root mean squared error (RMSE), coefficient of determination (R²) and spearman correlation ($\rho$) were noted. The latter one reflects the model’s ability to keep ranks between the objects. 

As it is difficult to determine the ground-truth Shore 00 value for fruits \cite{Yuan2017}, a test was set up to determine if predictions followed the ranks. Since the Shapiro-Wilk test revealed non-normality in some data groups, a non-parametric Wilcoxon rank-sum test ($\alpha=0.01$) was conducted to analyse whether, within one fruit sort, the median on the harder fruit was significantly higher than the softer fruit. As medians are compared, the interquartile range (IQR) will be given. In case of multiple comparisons (bananas and avocados), a Bonferroni-Holms correction was applied.

\subsubsection{\textbf{LLM Metrics}}

The LLM-as-a-judge is instructed to score the prompt from 1 to 5 based on three metrics: (i) Accuracy: are the objects, hardnesses and (relative) positions correctly described? (ii) Completeness: are all objects from the request mentioned and is info given if an object was not found? Is the ripeness interpreted for the correct cases? (iii) Clarity and Coherence: is the description understandable, concise and fluent?

\subsubsection{\textbf{Complete Framework Metrics}}

 Each scenario is executed 10 times, with all fruit equally present across the runs. In these scenarios, tactility values will not be tested again. Contrarily, an object-level (OL-SR) and scenario-level success rate (SL-SR) will be used. The OL-SR is defined as the average percentage of fruits the model was able to accurately identify, measure and communicate through the LLM. The SL-SR is more restrictive: it is defined as the percentage of times the total scenario was correctly executed and communicated. For instance, if 1 of 5 objects is mislocated, the OL-SR would be 4/5 while SL-SR 0. Finally, the latencies are reported as the average among the succeeded trials per scenario. A breakdown in latencies will reveal which steps take the longest.

\subsection{Main Results and Analysis}
\subsubsection{\textbf{Visual Servoing}}
In case the objects were identified, the independent t-test revealed that the confidence score of the YOLO model (0.921, 95\% CI:[0.886, 0.956]) was significantly higher (t=12.84, $p<0.01$, $n_1$=39, $n_2$=36) than the score for the GSAM model (0.645, [0.606, 0.685]). This is reflected in a SR of 0.9 for YOLO and 0.85 for GSAM. In the other cases, either the object was not found or another object had a higher confidence score than the target fruit.

However, in the correct cases, the segmentation score (IoU) of GSAM (0.942, [0.927, 0.956])] was significantly higher (t=9.01, $p<0.01$, $n_1$=39, $n_2$=36) than the score from YOLO (0.786, [0.752, 0.820]). Although the errors for YOLO (7.194mm, [5.534, 8.855]) were statistically not different (t=1.39, p=0.17, $n_1$=39, $n_2$=36) than GSAM (5.645mm, [4.314, 6.981]), only GSAM succeeded in having an error not statistically different than 5mm (t=1.35, p=0.19, $n=34$). In contrast, YOLO noted a statistically higher error than 5mm (t=2.80, $p<0.01$, $n$=36). This alignment with the tactile training setup makes the GSAM model more reliable for integration with the tactile perception model in TactEx.

\subsubsection{\textbf{Tactile Perception}}

\begin{table}[!t]
\centering
\captionsetup{font=footnotesize,labelsep=period}
\caption{RMSE, R² and $\rho$ for different models under various conditions. Lower RMSE, higher R² and Spearman $\rho$ are better. Results are shown after pretraining (80 epochs on online data) and fine-tuning (15 epochs on collected data). ResNet50-LSTM3: model employing a ResNet50 as CNN backbone and 3 LSTM layers. The baseline models use 2 contact images.}
\label{tab:pretfine}
\resizebox{1\linewidth}{!}{
\begin{tabular}{lccc|ccc}
\hline
\textbf{Model} & \multicolumn{3}{c}{\textbf{Pretraining}} & \multicolumn{3}{c}{\textbf{Fine-tuned}} \\
\cline{2-4} \cline{5-7}
 & RMSE & $R^2$ & $\rho$ & RMSE & $R^2$ & $\rho$ \\
\hline
\multicolumn{7}{c}{\hspace{80pt}\textit{Main Results}} \\
\hline
ResNet50-LSTM3 & 7.18 & 0.93 & 0.95 & 4.30 & 0.73 & 0.88 \\
Transformer & 6.87 & 0.93 & 0.94 & 6.23 & 0.44 & 0.77 \\
VGG-LSTM3 & 27.98 & -0.01 & -0.03 & 20.99 & -5.32 & 0.02 \\
\hline
\multicolumn{7}{c}{\hspace{80pt}\textit{Effect When Training on Half Scale}} \\
\hline
ResNet50-LSTM3 & 9.11 & 0.63 & 0.77 & 5.01 & 0.64 & 0.86 \\
Transformer & 9.03 & 0.65 & 0.81 & 4.33 & 0.73 & 0.89 \\
\hline
\multicolumn{7}{c}{\hspace{80pt}\textit{Effect of Different ResNet Backbone}} \\
\hline
ResNet34-LSTM3 & 6.83 & 0.94 & 0.96 & 6.50 & 0.41 & 0.89 \\
ResNet101-LSTM3 & 7.46 & 0.92 & 0.95 & 7.79 & 0.16 & 0.86 \\
\hline
\multicolumn{7}{c}{\hspace{80pt}\textit{Effect of More Contact Images (4)}} \\
\hline
ResNet50-LSTM3 & 7.13 & 0.93 & 0.96 & 8.80 & -0.11 & 0.73 \\
Transformer & 6.86 & 0.93 & 0.95 & 6.27 & 0.44 & 0.76 \\
\hline
\multicolumn{7}{c}{\hspace{80pt}\textit{Effect of LSTM Depth (1 layer)}} \\
\hline
ResNet50-LSTM1 & 7.43 & 0.93 & 0.95 & 10.63 & -0.65 & 0.78 \\
\hline
\multicolumn{7}{c}{\hspace{80pt}\textit{Effect of Direct Training}} \\
\hline
ResNet50-LSTM3 & - & - & - & 9.40 & 0.05 & 0.21 \\
\hline
\end{tabular}}
    \vspace{-0.4cm}
\end{table}

\begin{figure}[!h]
    \begin{subfigure}[b]{0.245\textwidth}
        \centering
        \includegraphics[width=\textwidth]{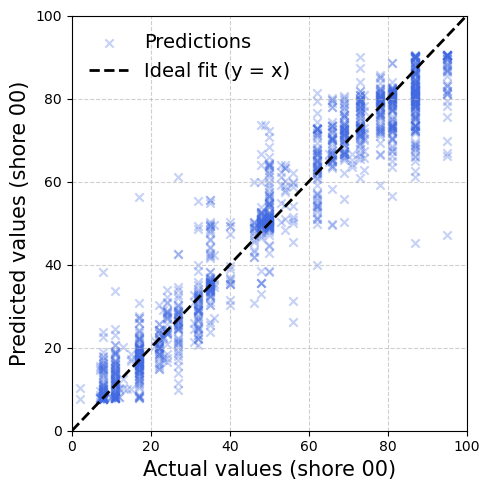}
        \caption{}
        \label{fig:sub12}
    \end{subfigure}
    \hspace{-9pt}
    \begin{subfigure}[b]{0.245\textwidth}
        \centering
        \includegraphics[width=\textwidth]{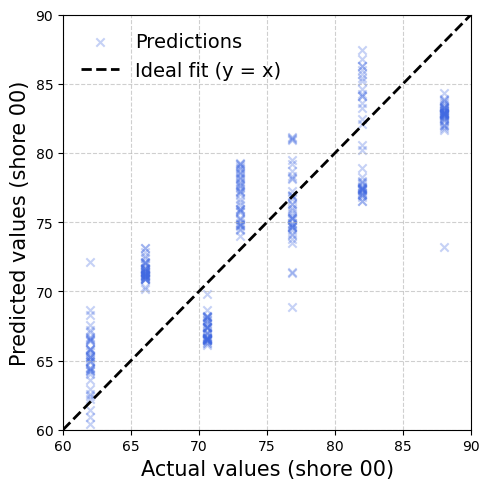}
        \caption{}
        \label{fig:sub22}
    \end{subfigure}
    \captionsetup{font=footnotesize,labelsep=period}
    \caption{Results of tactile predictions from the main ResNet50-LSTM3 model after pretraining (a) and fine-tuning (b). This is the model that will eventually be implemented within TactEx.}
    \label{fig:pretfine}
    \vspace{-0.2cm}
\end{figure}

The main results in Table \ref{tab:pretfine} reveal that the VGG16-LSTM was very prone to model collapsing, despite the extra measures taken. More specifically, the ResNet50-LSTM3 and Transformer baseline models in Table \ref{tab:pretfine} avoided model collapsing, thereby indicating that the applied measures were effective. The ResNet50-LSTM3 model achieved the best performance (lowest RMSE, highest R² and $\rho$) of the three models, both after pretraining and fine-tuning. 


\begin{table}[!t]
\centering
\captionsetup{font=footnotesize,labelsep=period}
\caption{Median and interquartile range of predictions (20 samples per individual fruit) from the full-range trained ResNet50-LSTM3 on different fruit pairs and trios. Hard, Medium, and Soft correspond to the empirical ripeness stages. Wilcoxon rank-sum test indicates whether the harder fruit is significantly ($p<0.01$) harder than the softer fruits.}
\label{tab:fruit_ranktests_resnet}
\begin{tabular}{lcccc}
\hline
\textbf{Condition} & Median & 25th & 75th & Wilcoxon \\
\hline
\multicolumn{5}{c}{\textit{Mango}} \\
\hline
Hard (1) & 79.47 & 72.89 & 84.59 & $u=343$, $p<0.01$ \\
Soft (0) & 67.75 & 65.60 & 72.78 &  \\
\hline
\multicolumn{5}{c}{\textit{Lime}} \\
\hline
Hard (1) & 64.13 & 63.78 & 64.64 & $u=285$, $p=0.011$ \\
Soft (0) & 63.84 & 63.73 & 63.93 &  \\
\hline
\multicolumn{5}{c}{\textit{Tomato}} \\
\hline
Hard (1) & 71.02 & 65.69 & 79.91 & $u=308$, $p<0.01$ \\
Soft (0) & 64.14 & 63.13 & 65.98 &  \\
\hline
\multicolumn{5}{c}{\textit{Banana}} \\
\hline
Hard (2)   & 72.63 & 67.50 & 82.53 & 2 vs 1: $u=288$, $p<0.01$ \\
Medium (1) & 66.87 & 66.31 & 67.62 & 1 vs 0: $u=362$, $p<0.01$ \\
Soft (0)   & 63.05 & 62.85 & 63.89 & 2 vs 0: $u=368$, $p<0.01$ \\
\hline
\multicolumn{5}{c}{\textit{Avocado}} \\
\hline
Hard (2)   & 65.25 & 64.02 & 65.92 & 2 vs 1: $u=299$, $p<0.01$ \\
Medium (1) & 63.54 & 63.39 & 64.12 & 1 vs 0: $u=359$, $p<0.01$ \\
Soft (0)   & 61.73 & 60.97 & 62.15 & 2 vs 0: $u=373$, $p<0.01$ \\
\hline
\end{tabular}
\vspace{-0.4cm}
\end{table}

The ResNet50-LSTM3 results after pretraining and fine-tuning are visualized in Fig. \ref{fig:pretfine}. Due to its superior performance among the primary results, the ResNet50-LSTM3 was selected for validation in the fruit ranking scenario.

The validation results are reported in Table \ref{tab:fruit_ranktests_resnet}. The Wilcoxon rank-sum tests showed that all comparisons were statistically significant ($p<0.01)$. This demonstrates that the model can correctly interpret which fruit is harder, thereby mimicking human touch. This statistical significance is something not earlier discovered in literature \cite{Yuan2024, Liao2025}. We dedicate this result to the change in backbone, training strategy and study of the optimal architecture, as explained in the ablation study (section \ref{sec:abl}).

\subsubsection{\textbf{LLM}}
The LLM-as-a-judge scores on 5 reveal a solid performance of the LLM answer: the accuracy was 4.19 $\pm$ 0.59, completeness 4.94 $\pm$ 0.24 and conciseness and clarity 4.92 $\pm$ 0.44.

\subsubsection{\textbf{Integrated Framework}}

\begin{figure*}[ht!]
    \centering
    \captionsetup{font=footnotesize,labelsep=period}
    \includegraphics[width=1\linewidth]{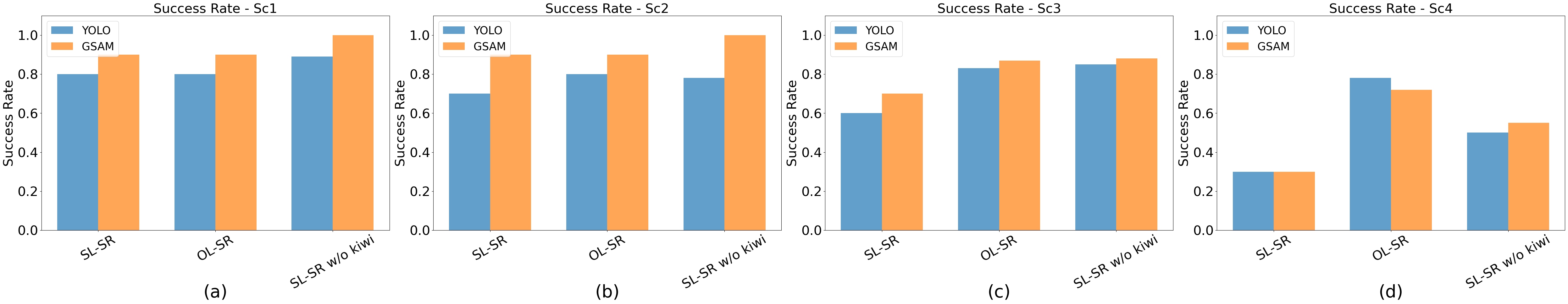}
    \caption{Success rates of the TactEx framework across four interaction scenarios of increasing complexity (Sc1-Sc4), as defined by Table \ref{tab:complexity}. SL-SR: Scenario-Level Success Rate, OL-SR: Object-Level Success Rate, w/o: without, Sc: Senario.}

        \vspace{-0.2cm}
    \label{fig:sr}
\end{figure*}

\begin{figure*}[h!]
    \centering
    \captionsetup{font=footnotesize,labelsep=period}
    \includegraphics[width=1\linewidth]{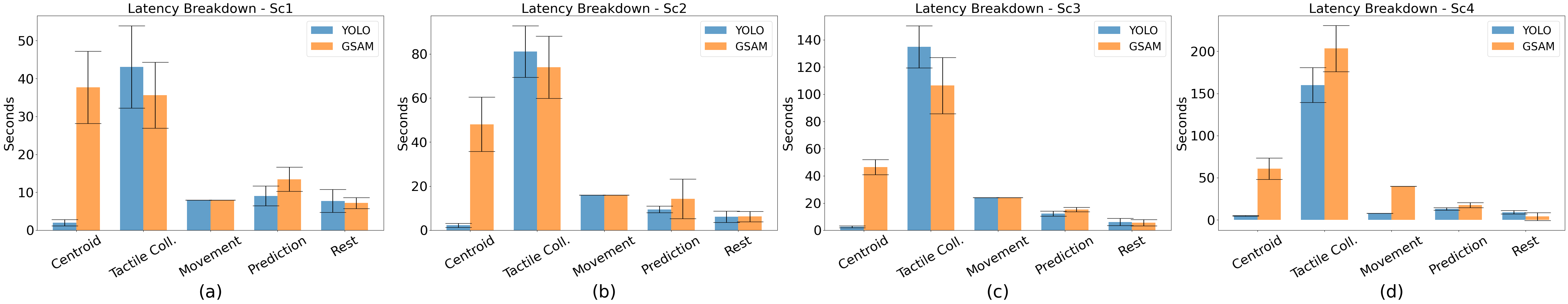}
    \caption{Latency breakdown across the different scenarios (Sc1–Sc4), as defined by Table \ref{tab:complexity}. Tactile collection takes the longest in both models, while GSAM also suffers from heavy computational delays during centroid computation. Sc: Scenario, Coll: Collection.}
    \label{fig:lat}
    \vspace{-0.4cm}
\end{figure*}

The success rate are visualized in Fig. \ref{fig:sr}. The mismatch between SL-SR and OL-SR in Fig. \ref{fig:sr} reveals that both models have problems with certain fruits, more specifically a kiwi. This strongly impacts the SL-SR. The effect was further amplified in more complex scenarios where kiwis appeared more frequently. Indeed, discarding the kiwi cases improved the performance dramatically. For YOLO, this likely reflects the low amount of kiwi images (15) and too little training (100 epochs). We encourage future researchers to include at least 25 images per class to avoid these class imbalance issues and train for 150 epochs. 

The decrease across higher complexity is consistent with literature \cite{Verbaan2024}. The SL-SR in Fig. \ref{fig:sr} demonstrates that when the fruits are explicitly stated, strong performance is achieved. This shows the model’s ability to select the hardest fruit on the table. In the case of a banana, lime and lemon, the LLM interprets these hardnesses as ripe or unripe with high accuracy, making it practical in household applications. 
In the fourth scenario however, the SL-SR drops significantly, even without kiwis. This is explained by the fact that the prompt did not mention the fruits explicitly. In contrary, a list of 20 fruits and vegetables was prompted in GSAM. However, without defined fruits of interest, we cannot use the NLP logic to filter out mistakes. This ultimately leads to more missdetections, with GSAM more affected than YOLO, as shown by the OL-SR. The latter is explained by the lower confidence score reported earlier, stemming from the fact that it is not specifically trained on these targets, unlike YOLO.

In general, however, GSAM outperforms YOLO by 10 percentage point. This is explained by the lower error and better segmentation reported earlier, which makes tactile prediction more reliable as it adheres more to the tactile training setup. This further illustrates the need for longer training and more data in YOLO. In contrast, GSAM is more versatile and deployable into new applications compared to YOLO as it does not require any training. The combination of GSAM and tactile perception makes TactEx an optimal solution for household, agricultural, and industrial applications.


\subsection{Ablation Study}
\label{sec:abl}
\subsubsection{Ablation Results}
We conducted an ablation study to assess key design choices: number of contact images (2 vs. 4), LSTM depth (1 vs. 3), and CNN backbone (ResNet34, ResNet50, ResNet101). We also compared direct training on the collected samples against the pretraining strategy, and examined pretraining restricted to the upper half of the Shore 00 scale ($>39$) to test its effect after fine-tuning. The results of these variants are summarized in Table \ref{tab:pretfine}.

\subsubsection{Analysis}
\label{sec:Analysis}
Table \ref{tab:pretfine} suggests a new effective training strategy for future hardness estimation models. Four findings support this statement. Firstly, the directly trained ResNet50-LSTM3 performed poorly (RMSE 9.40) compared to the pretraining strategy (RMSE 4.30). Second, while the pretrained main models achieve slightly higher RMSE than related works (6.87 vs 5.18), the R² values are consistent with state-of-the-art models \cite{Yuan2017, Nam2024}. Third, the fine-tuned models demonstrate that even with a low amount of collected data the model seems to generalize well to a new robot setup. Indeed, the spearman correlations remain high, with only minor inconsistencies visible in Fig. \ref{fig:pretfine}. This suggests that the model is great at ranking hardness values, with minimal deviations from the true values (RMSE 4.30). Fourth and finally, none of the visualized models in Fig. \ref{fig:pretfine} reveal systematic degradation for harder objects, suggesting that the Gelsight sensor is better suited than the TacTip used by Nam et al. (2024) \cite{Nam2024}.

 With regard to the optimal architecture, the ablation results were influenced by the glass pouch object. Nevertheless, we consider the conclusions from the ablation study as valid, as they are mostly based on the spearman correlation. Three design choices emerged as most effective: (1) four contact images do not improve the model after pretraining and fine-tuning suggesting that earlier research by Nam et al. (2024) was right in that the first and last contact image are most crucial \cite{Nam2024}, (2) three LSTM layers better capture the subtle variation between contact images and (3) while we believe the other ResNet backbones also work for our use case ($\rho\ge0.80$), the ResNet50 seems to balance complexity and robustness more optimally (lower RMSE and higher R²). 
 
Considering the pretraining range, results are mixed: the Transformer model benefited from pretraining on half the range ($\rho = 0.89$ vs. $0.77$), whereas the ResNet50-LSTM3 model did not ($\rho = 0.86$ vs. $0.88$). This is likely due to the Transformer’s attention mechanism, which may emphasize localized patterns when the training range is narrower. Therefore, our recommendation is as follows: if the application of interest has a predefined target range, it is worthy to investigate whether pretraining on that range may help.

\section{Conclusion and Future Work}

We presented \textbf{TactEx}, an explainable multimodal robotic framework that integrates vision, language grounding, tactile exploration, and robotic control to estimate and communicate object hardness in natural language. Across the evaluated scenarios, the ResNet50--LSTM3 architecture demonstrated reliable tactile hardness regression, preserving expected hardness rankings with statistically significant differences under non-parametric testing. These results support the feasibility of interactive, language-driven fruit ripeness assessment using vision-guided tactile sensing.
The modular integration of GSAM-based object grounding, tactile inference, and LLM-based explanation enables component-level replacement and adaptation to new sensors, object categories, and tasks without redesigning the overall system.

Despite these contributions, the current system operates with latency due to sequential perception and execution. Future work should explore tighter perception–action coupling, including preprocessing visual scenes prior to user prompts, incremental scene updating, and closed-loop tactile exploration policies that adapt probing online. 

\bibliographystyle{ieeetr}
\bibliography{sample}

@misc{huang2025tactilevlaunlockingvisionlanguageactionmodels,
      title={Tactile-VLA: Unlocking Vision-Language-Action Model's Physical Knowledge for Tactile Generalization}, 
      author={Jialei Huang and Shuo Wang and Fanqi Lin and Yihang Hu and Chuan Wen and Yang Gao},
      year={2025},
      eprint={2507.09160},
      archivePrefix={arXiv},
      primaryClass={cs.RO},
      url={https://arxiv.org/abs/2507.09160}, 
}

@inproceedings{li2018slip,
  title={Slip detection with combined tactile and visual information},
  author={Li, Jianhua and Dong, Siyuan and Adelson, Edward},
  booktitle={2018 IEEE International Conference on Robotics and Automation (ICRA)},
  pages={7772--7777},
  year={2018},
  organization={IEEE}
}

@article{SETCHI20203057,
title = {Explainable Robotics in Human-Robot Interactions},
journal = {Procedia Computer Science},
volume = {176},
pages = {3057-3066},
year = {2020},
note = {Knowledge-Based and Intelligent Information \& Engineering Systems: Proceedings of the 24th International Conference KES2020},
issn = {1877-0509},
doi = {https://doi.org/10.1016/j.procs.2020.09.198},
url = {https://www.sciencedirect.com/science/article/pii/S1877050920321001},
author = {Rossitza Setchi and Maryam Banitalebi Dehkordi and Juwairiya Siraj Khan}
}

@article{adebayo,
author = {Adebayo, Abiodun and Ajayi, Olanrewaju and Chukwurah, Naomi and Pub, Anfo},
year = {2024},
month = {03},
pages = {26-32},
title = {Explainable AI in Robotics: A Critical Review and Implementation Strategies for Transparent Decision-Making},
volume = {05},
journal = {Journal Of Multidisciplinary Research}
}

@misc{zhang2025vtlavisiontactilelanguageactionmodelpreference,
      title={VTLA: Vision-Tactile-Language-Action Model with Preference Learning for Insertion Manipulation}, 
      author={Chaofan Zhang and Peng Hao and Xiaoge Cao and Xiaoshuai Hao and Shaowei Cui and Shuo Wang},
      year={2025},
      eprint={2505.09577},
      archivePrefix={arXiv},
      primaryClass={cs.RO},
      url={https://arxiv.org/abs/2505.09577}, 
}

@misc{yu2025forcevlaenhancingvlamodels,
      title={ForceVLA: Enhancing VLA Models with a Force-aware MoE for Contact-rich Manipulation}, 
      author={Jiawen Yu and Hairuo Liu and Qiaojun Yu and Jieji Ren and Ce Hao and Haitong Ding and Guangyu Huang and Guofan Huang and Yan Song and Panpan Cai and Cewu Lu and Wenqiang Zhang},
      year={2025},
      eprint={2505.22159},
      archivePrefix={arXiv},
      primaryClass={cs.RO},
      url={https://arxiv.org/abs/2505.22159}, 
}

@misc{Chen2025,
      title={Investigating Active Sampling for Hardness Classification with Vision-Based Tactile Sensors}, 
      author={Junyi Chen and Alap Kshirsagar and Frederik Heller and Mario Gómez Andreu and Boris Belousov and Tim Schneider and Lisa P. Y. Lin and Katja Doerschner and Knut Drewing and Jan Peters},
      year={2025},
      eprint={2505.13231},
      archivePrefix={arXiv},
      primaryClass={cs.RO},
      url={https://arxiv.org/abs/2505.13231}, 
}

@inproceedings{Nam2024,
   author = {Saekwang Nam and Toby Jack and Loong Yi Lee and Nathan F. Lepora},
   doi = {10.1109/RoboSoft60065.2024.10521971},
   isbn = {9798350381818},
   booktitle = {2024 IEEE 7th International Conference on Soft Robotics, RoboSoft 2024},
   pages = {121-126},
   publisher = {Institute of Electrical and Electronics Engineers Inc.},
   title = {Softness Prediction with a Soft Biomimetic Optical Tactile Sensor},
   year = {2024}
}

@INPROCEEDINGS{Yuan2024,
  author={Yuan, Wenzhen and Srinivasan, Mandayam A. and Adelson, Edward H.},
  booktitle={2016 IEEE/RSJ International Conference on Intelligent Robots and Systems (IROS)}, 
  title={Estimating object hardness with a GelSight touch sensor}, 
  year={2016},
  volume={},
  number={},
  pages={208-215},
  doi={10.1109/IROS.2016.7759057}}

@misc{Calandra2017,
      title={The Feeling of Success: Does Touch Sensing Help Predict Grasp Outcomes?}, 
      author={Roberto Calandra and Andrew Owens and Manu Upadhyaya and Wenzhen Yuan and Justin Lin and Edward H. Adelson and Sergey Levine},
      year={2025},
      eprint={1710.05512},
      archivePrefix={arXiv},
      primaryClass={cs.RO},
      url={https://arxiv.org/abs/1710.05512}, 
}

@misc{Gao2023,
      title={Visuo-Tactile-Based Slip Detection Using A Multi-Scale Temporal Convolution Network}, 
      author={Junli Gao and Zhaoji Huang and Zhaonian Tang and Haitao Song and Wenyu Liang},
      year={2023},
      eprint={2302.13564},
      archivePrefix={arXiv},
      primaryClass={cs.RO},
      url={https://arxiv.org/abs/2302.13564}, 
}

@misc{Yu2024,
      title={Octopi: Object Property Reasoning with Large Tactile-Language Models}, 
      author={Samson Yu and Kelvin Lin and Anxing Xiao and Jiafei Duan and Harold Soh},
      year={2024},
      eprint={2405.02794},
      archivePrefix={arXiv},
      primaryClass={cs.RO},
      url={https://arxiv.org/abs/2405.02794}, 
}

@inproceedings{Fu2024,
   author = {Letian Fu and Gaurav Datta and Huang Huang and William Chung-Ho Panitch and Jaimyn Drake and Joseph Ortiz and Mustafa Mukadam and Mike Lambeta and Roberto Calandra and Ken Goldberg},
   city = {Vienna, Austria},
   institution = {UC Berkeley, META AI, TU Tresden},
   booktitle = {Proceedings of the 41 st International Conference on Machine Learning},
   title = {A Touch, Vision, and Language Dataset for Multimodal Alignment},
   year = {2024}
}

@inproceedings{Ueda2024,
   author = {Shiori Ueda and Atsushi Hashimoto and Masashi Hamaya and Kazutoshi Tanaka and Hideo Saito},
   booktitle = {International Conference on Intelligent Robots and Systems},
   month = {9},
   publisher = {IEEE/RSJ},
   title = {Visuo-Tactile Zero-Shot Object Recognition with Vision-Language Model},
   url = {http://arxiv.org/abs/2409.09276},
   year = {2024}
}

@inproceedings{Yuan2017,
   title={Shape-independent hardness estimation using deep learning and a GelSight tactile sensor},
   url={http://dx.doi.org/10.1109/ICRA.2017.7989116},
   DOI={10.1109/icra.2017.7989116},
   booktitle={2017 IEEE International Conference on Robotics and Automation (ICRA)},
   publisher={IEEE},
   author={Yuan, Wenzhen and Zhu, Chenzhuo and Owens, Andrew and Srinivasan, Mandayam A. and Adelson, Edward H.},
   year={2017},
   month=may, pages={951–958} }

@misc{Liao2025,
      title={Quantitative Hardness Assessment with Vision-based Tactile Sensing for Fruit Classification and Grasping}, 
      author={Zhongyuan Liao and Yipai Du and Jianghua Duan and Haobo Liang and Michael Yu Wang},
      year={2025},
      eprint={2505.05725},
      archivePrefix={arXiv},
      primaryClass={cs.RO},
      url={https://arxiv.org/abs/2505.05725}, 
}

@article{zhang2025design,
  title={Design and benchmarking of a multimodality sensor for robotic manipulation with GAN-based cross-modality interpretation},
  author={Zhang, Dandan and Fan, Wen and Lin, Jialin and Li, Haoran and Cong, Qingzheng and Liu, Weiru and Lepora, Nathan F and Luo, Shan},
  journal={IEEE Transactions on Robotics},
  volume={41},
  pages={1278--1295},
  year={2025},
  publisher={IEEE}
}

@article{su2012use,
  title={Use of tactile feedback to control exploratory movements to characterize object compliance},
  author={Su, Zhe and Fishel, Jeremy A and Yamamoto, Tomonori and Loeb, Gerald E},
  journal={Frontiers in neurorobotics},
  volume={6},
  pages={7},
  year={2012},
  publisher={Frontiers Media SA}
}

@article{fan2025crystaltac,
  title={CrystalTac: Vision-based tactile sensor family fabricated via rapid monolithic manufacturing},
  author={Fan, Wen and Li, Haoran and Zhang, Dandan},
  journal={Cyborg and Bionic Systems},
  volume={6},
  pages={0231},
  year={2025},
  publisher={AAAS}
}

@article{Zhang2024,
   author = {Xuyang Zhang and Tianqi Yang and Dandan Zhang and Nathan F. Lepora},
   doi = {10.1109/JSEN.2024.3471812},
   issn = {15581748},
   journal = {IEEE Sensors Journal},
   publisher = {Institute of Electrical and Electronics Engineers Inc.},
   title = {TacPalm: A Soft Gripper with a Biomimetic Optical Tactile Palm for Stable Precise Grasping},
   year = {2024}
}

@article{zhang2021explainable,
  title={Explainable hierarchical imitation learning for robotic drink pouring},
  author={Zhang, Dandan and Li, Qiang and Zheng, Yu and Wei, Lei and Zhang, Dongsheng and Zhang, Zhengyou},
  journal={IEEE Transactions on Automation Science and Engineering},
  volume={19},
  number={4},
  pages={3871--3887},
  year={2021},
  publisher={IEEE}
}

@inproceedings{fan2022graph,
  title={Graph neural networks for interpretable tactile sensing},
  author={Fan, Wen and Bo, Hongbo and Lin, Yijiong and Xing, Yifan and Liu, Weiru and Lepora, Nathan and Zhang, Dandan},
  booktitle={2022 27th International Conference on Automation and Computing (ICAC)},
  pages={1--6},
  year={2022},
  organization={IEEE}
}

@INPROCEEDINGS{10160288,
  author={Fan, Wen and Yang, Max and Xing, Yifan and Lepora, Nathan F. and Zhang, Dandan},
  booktitle={2023 IEEE International Conference on Robotics and Automation (ICRA)}, 
  title={Tac-VGNN: A Voronoi Graph Neural Network for Pose-Based Tactile Servoing}, 
  year={2023},
  volume={},
  number={},
  pages={10373-10379},
  doi={10.1109/ICRA48891.2023.10160288}}

@techReport{TianheRen2024,
   author = {Tianhe Ren and Shilong Liu and Ailing Zeng and Jin Ling and He Cao and Kunchang Li and Jiayu Chen and Xinyu Huang and Feng Yan and YUkang Chen},
   institution = {International Digital Economy Academy (IDEA) \& Community},
   month = {1},
   title = {Grounded SAM: Assembling Open-World Models for Diverse Visual Tasks International Digital Economy Academy (IDEA) \& Community},
   url = {https://github.com/IDEA-Research/Grounded-Segment-Anything},
   year = {2024}
}

@article{Bai2020,
   author = {Qiang Bai and Shaobo Li and Jing Yang and Qisong Song and Zhiang Li and Xingxing Zhang},
   doi = {10.1109/ACCESS.2020.3028740},
   issn = {21693536},
   journal = {IEEE Access},
   pages = {181855-181879},
   publisher = {Institute of Electrical and Electronics Engineers Inc.},
   title = {Object detection recognition and robot grasping based on machine learning: A survey},
   volume = {8},
   year = {2020}
}

@phdthesis{Verbaan2024,
   author = {Leonoor Verbaan and Yke Bauke Eisma and Remco van Leeuwen},
   city = {Delft},
   school = {Delft University of Technology},
   month = {8},
   title = {Perception and Control with Large Language Models in Robotic Manipulation Developing and assessing an integrated Large Language Model System on environmental and task complexity},
   year = {2024}
}

@misc{Gu2025,
      title={A Survey on LLM-as-a-Judge}, 
      author={Jiawei Gu and Xuhui Jiang and Zhichao Shi and Hexiang Tan and Xuehao Zhai and Chengjin Xu and Wei Li and Yinghan Shen and Shengjie Ma and Honghao Liu and Saizhuo Wang and Kun Zhang and Yuanzhuo Wang and Wen Gao and Lionel Ni and Jian Guo},
      year={2025},
      eprint={2411.15594},
      archivePrefix={arXiv},
      primaryClass={cs.CL},
      url={https://arxiv.org/abs/2411.15594}, 
}

@inproceedings{Zhao2023,
   author = {Xufeng Zhao and Mengdi Li and Cornelius Weber and Muhammad Burhan Hafez and Stefan Wermter},
   doi = {10.1109/IROS55552.2023.10342363},
   isbn = {9781665491907},
   issn = {21530866},
   booktitle = {IEEE International Conference on Intelligent Robots and Systems},
   pages = {3590-3596},
   publisher = {Institute of Electrical and Electronics Engineers Inc.},
   title = {Chat with the Environment: Interactive Multimodal Perception Using Large Language Models},
   year = {2023}
}

@Article{s20133796,
AUTHOR = {Lee, Jong-il and Lee, Suwoong and Oh, Hyun-Min and Cho, Bo Ram and Seo, Kap-Ho and Kim, Min Young},
TITLE = {3D Contact Position Estimation of Image-Based Areal Soft Tactile Sensor with Printed Array Markers and Image Sensors},
JOURNAL = {Sensors},
VOLUME = {20},
YEAR = {2020},
NUMBER = {13},
ARTICLE-NUMBER = {3796},
URL = {https://www.mdpi.com/1424-8220/20/13/3796},
PubMedID = {32645894},
ISSN = {1424-8220},
DOI = {10.3390/s20133796}
}

@misc{Guo2025,
      title={Robotic Perception with a Large Tactile-Vision-Language Model for Physical Property Inference}, 
      author={Zexiang Guo and Hengxiang Chen and Xinheng Mai and Qiusang Qiu and Gan Ma and Zhanat Kappassov and Qiang Li and Nutan Chen},
      year={2025},
      eprint={2506.19303},
      archivePrefix={arXiv},
      primaryClass={cs.RO},
      url={https://arxiv.org/abs/2506.19303}, 
}

\end{document}